# Classification of Driver Behaviour Using External Observation Techniques for Autonomous Vehicles

Ian Nell
*Faculty of Engineering*
*Atlantic Technological University*
Sligo, Ireland
S00166888@atu.ie

Shane Gilroy*
*Faculty of Engineering*
*Atlantic Technological University*
Sligo, Ireland
shane.gilroy@atu.ie
*Corresponding Author

***Abstract*—Road traffic accidents remain a significant global concern, with human error, particularly distracted and impaired driving, among the leading causes. This study introduces a novel driver behaviour classification system that uses external observation techniques to detect indicators of distraction and impairment. The proposed framework employs advanced computer vision methodologies, including real-time object tracking, lateral displacement analysis, and lane position monitoring. The system identifies unsafe driving behaviours such as excessive lateral movement and erratic trajectory patterns by implementing the YOLO object detection model and custom lane estimation algorithms. Unlike systems reliant on inter-vehicular communication, this vision-based approach enables behavioural analysis of non-connected vehicles. Experimental evaluations on diverse video datasets demonstrate the framework's reliability and adaptability across varying road and environmental conditions.**

## I. Introduction

Road traffic accidents remain a significant global concern, with human error, particularly distracted and impaired driving, among the leading causes [1]. According to the World Health Organization's Global Status Report on Road Safety 2023, road traffic deaths reached an estimated 1.19 million people in 2021, with speeding, drunk driving, distracted driving, and unsafe vehicles being primary contributors [1]. These preventable actions underscore the urgent need for innovative solutions to enhance road safety. The report also highlights the importance of addressing vulnerable road users, such as pedestrians and cyclists, who are disproportionately affected by unsafe driving behaviours [1]. Advancements in Intelligent Transport Systems (ITS), Advanced Driver Assistance Systems (ADAS), and Artificial Intelligence (AI) have provided opportunities to address these challenges. ADAS features, such as lane departure warnings, emergency braking, and adaptive speed control, have demonstrated their ability to reduce accidents caused by human error by up to one-third [2]. Machine learning (ML) techniques have been widely used to study driver behaviour, such as tracking eye movements, detecting signs of distraction, and identifying fatigue [2], [6]. These technologies have shown high accuracy in detecting unsafe driving behaviours, particularly when combined with diverse and realistic datasets [2]. Despite these advancements, many vehicles on public roads fall within SAE Levels 0 to 2, lacking advanced connectivity features like Vehicle-to-Vehicle (V2V) communication [3]. This presents challenges for autonomous or semi-autonomous systems to detect and respond to unsafe driving behaviours in mixed-traffic conditions. The absence of inter-vehicular communication necessitates the development of systems capable of externally monitoring driver behaviour to infer cognitive or physiological states, such as distraction or impairment, based solely on observable driving patterns [2], [4]. For example, lateral movement, oscillatory behaviour, and erratic lane positioning can indicate unsafe driving. These patterns can be analyzed using advanced computer vision techniques to classify driver behaviour in real-time [5].

This research proposes a vision-based approach to classify driver behaviour using external observation techniques. By leveraging advanced computer vision methodologies, including real-time object tracking, lateral displacement analysis, and lane position monitoring, the system aims to detect unsafe driving patterns without relying on inter-vehicular communication. Integrating the YOLO object detection model and custom lane estimation algorithms enables the identification of aberrant driving behaviours indicative of distraction or impairment [5]. The research addresses the critical need for systems to enhance road safety during the transitional phase where manually driven and autonomous vehicles coexist. By focusing on externally observable driving patterns, such as lateral movement and oscillatory behaviour, the proposed framework offers a scalable solution to improve hazard detection in environments where direct vehicle communication is absent [2], [4]. Experimental evaluations conducted on diverse video datasets from real-world driving scenarios highlight the framework's capability to detect unsafe behaviour reliably [5]. The proposed methodology demonstrates adaptability and robustness despite inherent challenges, such as faded or absent lane markings, complex road geometries, environmental variability, and occlusions [5]. By addressing the limitations of current autonomous vehicle systems, this research contributes to the broader vision of safe, intelligent, and fully autonomous transportation systems [2], [4].

## II. Related Work

Driver behaviour monitoring systems use various data sources to detect unsafe driving patterns, such as distraction, fatigue, and impairment. Kotseruba and Tsotsos [2] reviewed algorithms and datasets for vision-based assistive and automated driving systems. Their findings highlighted the importance of visual attention in driving and identified key challenges, such as limited scope in current attention models and the need for diverse datasets. ADAS features, including lane departure warnings, adaptive speed control, and emergency braking, have demonstrated their ability to reduce accidents caused by human error by up to one-third [2]. Zaidan et al. [4] emphasized the critical role of understanding driver behaviour in developing ITS and ADAS. The authors categorized driver behaviour analysis techniques into onboard system analysis, smartphone-based analysis, and predetermined data analysis. Onboard systems use embedded sensors like LiDAR and accelerometers to measure steering and braking parameters, while smartphone-based systems offer cost-effective real-time feedback. Predetermined data analysis leverages large datasets to develop predictive algorithms and models. The car-following model is fundamental in understanding the interaction between autonomous vehicles (AVs) and human-driven vehicles (HDVs) in traffic scenarios. It is critical to ensure safe and efficient driving by maintaining appropriate headway distances and controlling vehicle speed and acceleration. This model is particularly relevant in mixed-traffic environments where AVs and HDVs coexist. Car-following models simulate the behaviour of a trailing vehicle (referred to as the "ego vehicle") as it follows a lead vehicle. These models aim to maintain a safe driving distance by dynamically adjusting the speed and acceleration of the trailing vehicle based on the movements of the lead vehicle. The primary goal is to prevent collisions while optimizing traffic flow [6]. For AVs, car-following models are essential for: (a) Longitudinal Control: Ensuring smooth acceleration and deceleration to maintain safe distances. (b) Lane-Keeping: Adjusting lateral positioning to stay within lane boundaries. (c) Headway Distance Maintenance: Calculating the optimal distance between vehicles to avoid abrupt braking or acceleration [6]. There are three primary categories of Car-Following Models, *Analytical Models* that use mathematical equations to predict vehicle behaviour based on speed, acceleration, and headway distance. *Rule-Based Models* that rely on predefined rules to determine vehicle actions, such as maintaining a minimum safe distance and *Simulation-Based Models* that simulate real-world driving scenarios to test AV behaviour under diverse conditions [6].

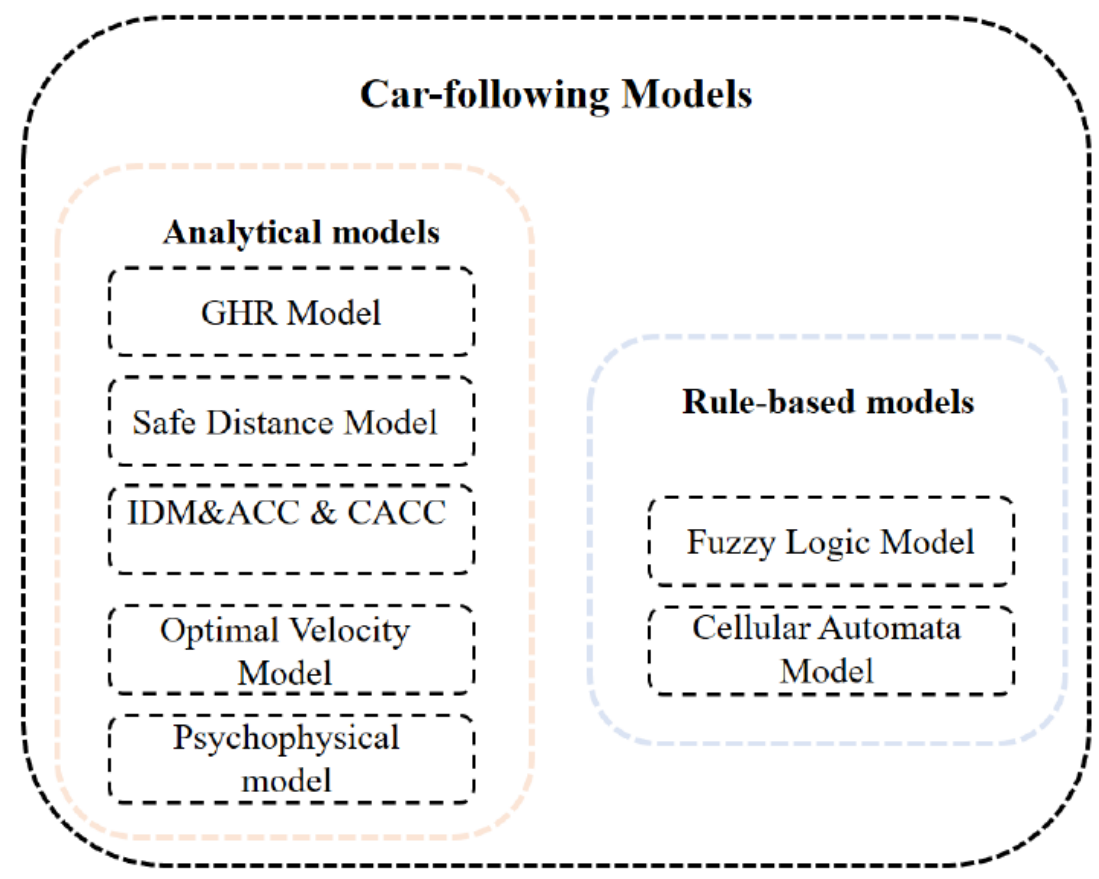

Fig. 1. Car-Following Models

Car-following models face several challenges such as: Mixed-Traffic Conditions: Interactions between AVs and HDVs can be unpredictable, requiring adaptive algorithms. Environmental Variability: Road conditions, weather, and traffic density can impact model performance. Human Behaviour: HDVs may exhibit erratic or aggressive driving patterns, complicating the predictive capabilities of car-following models [6].

Machine Learning techniques have been widely applied to study driver behaviour. CNNs are commonly used for gaze estimation and object detection, while RNNs are employed for distraction detection and sequence analysis [2], [7]. Koay et al. [7] reviewed ML techniques tailored to distraction detection, including traditional ML algorithms, CNNs, and RNNs. The study summarized publicly available datasets and highlighted the importance of multi-modal approaches for improving detection accuracy. Shahverdy et al. [8] proposed a deep learning method for classifying driver behaviours into five categories: standard, aggressive, distracted, drowsy, and drunk. Their experiments involved collecting driving data under real-world conditions and converting signals into images using recurrence plot techniques. Results demonstrated high accuracy (99.98%) in detecting unsafe driving behaviours, with CNN configurations optimized for embedded applications.

Sensor fusion techniques integrate data from multiple sources, such as cameras, LiDAR, RADAR, and physiological sensors, to enhance the capabilities of autonomous vehicles and ADAS. Hafeez et al. [9] reviewed existing literature on sensor fusion technologies and highlighted their advantages in improving spatial awareness and pedestrian intention prediction. For example, combining RGB images with LiDAR point cloud data improves object detection, trajectory planning, and collision avoidance [9]. Chen and Chen [10] introduced the Driver Behaviour Monitoring and Warning (DBMW) framework, which integrates onboard image sensors and wearable devices to monitor driver behaviour and vehicle lane deviation. Their system uses Power Spectral Density (PSD) classification to measure lane deviation and the Longest Common Subsequence (LCS) algorithm to detect head motion anomalies. The framework demonstrated high effectiveness in spotting dangerous driving behaviours and issuing timely warnings.

Naturalistic Driving Studies (NDS) provide valuable insights into driver behaviour by observing real-world driving scenarios. Large-scale studies, such as the 100-Car NDS and the Second Strategic Highway Research Program (SHRP 2), have contributed to understanding factors influencing road accidents [11]. Wang et al. [8] discussed the challenges of analyzing large NDS datasets and the potential of computer vision and ML techniques to automate data annotation and driver behaviour classification. Ahmed et al. [12] analyzed NDS data to study driver behaviour under varying weather and traffic conditions. Their findings revealed that drivers adjust their behaviour in extreme weather conditions, such as maintaining greater distances between vehicles to compensate for reduced visibility. These insights are crucial for developing safety countermeasures and improving traffic operations.

Despite significant advancements, several limitations persist in current driver behaviour monitoring systems. Many existing approaches rely on in-vehicle sensors, such as infrared cameras, steering wheel movement trackers, or smartphone-based telemetry. These are constrained in real-world scenarios where vehicles lack connectivity features like V2V or V2X communication [2], [4]. Additionally, the overreliance on narrowly defined attention models and limited datasets compromises the ability to generalize across diverse road conditions, lighting environments, and cultural driving norms [7]. Researchers advocate for multidisciplinary frameworks that combine ML, behavioural psychology, computer vision, and human factors engineering to address these gaps. Expanding datasets to reflect diverse driving scenarios and integrating multi-modal sensors, such as LiDAR, GPS, and IMUs, can significantly enhance detection accuracy and system reliability [9], [10]. Establishing standardized performance metrics for evaluating driver monitoring systems is essential for ensuring application consistency and reliability [11].

Car-following models are a cornerstone of autonomous-vehicle technology, enabling safe and efficient navigation in mixed-traffic environments. Their integration into AV systems ensures smoother transitions between human-driven and fully autonomous vehicles, advancing the broader vision of intelligent transportation systems. Alongside these models, recent breakthroughs in driver-behaviour monitoring, machine-learning techniques, sensor-fusion technologies, and naturalistic driving studies have shown strong potential to improve road safety and accelerate AV development. However, key challenges remain—notably the lack of inter-vehicular communication frameworks and the need for richer, more diverse datasets—to achieve reliable and scalable solutions. The vision-based approach proposed in this research seeks to bridge these gaps by focusing on externally observable driving patterns, offering a promising path toward safer roads throughout the transition to fully autonomous vehicles.

## III. METHODOLOGY

This research proposes a vision-based driver behaviour classification system to detect distracted and impaired driving using external observation techniques. The methodology is structured as follows:

### A. Research Design

The system analyzes lateral vehicle movements and lane positioning to identify unsafe driving behaviours. A monocular camera was selected as the primary sensor due to its cost-effectiveness and ease of deployment, eliminating the need for inter-vehicular communication systems [13], [15].

### B. Data Collection

Video recordings were captured using a Garmin 55 dashcam mounted on the windshield of the ego vehicle. The recordings include diverse road types under varying lighting and environmental conditions, including highways, local roads, and back streets. Controlled simulations of distracted and impaired driving were conducted to evaluate the system's performance.

### C. Object Detection

The YOLOv8n (You Only Look Once) object detection model was employed to identify vehicles and other road users in real-time. YOLOv8n was selected for its lightweight architecture, high inference speed (≈ 2–4 ms/frame), and competitive accuracy (≈ 37% mPA at IoU = 0.5) [15]. The model was trained using the Microsoft COCO dataset, which provides robust annotations for object detection tasks [17].

### D. Lane Detection

Lane markings were detected using a multi-stage image preprocessing pipeline:

1. *Grayscale Conversion*: Simplifies image processing by reducing colour complexity.
2. *Gaussian Blurring:* Suppresses high-frequency noise for better edge detection.
3. *Adaptive Thresholding:* Handles variable lighting conditions to enhance lane markings.
4. *Canny Edge Detection:* Extracts prominent structural features from the image [18]. Detected lane lines were fitted using RANSAC regression to minimize the influence of noise and outliers. Polynomial curves were used to estimate lane boundaries and the lane centre [19].

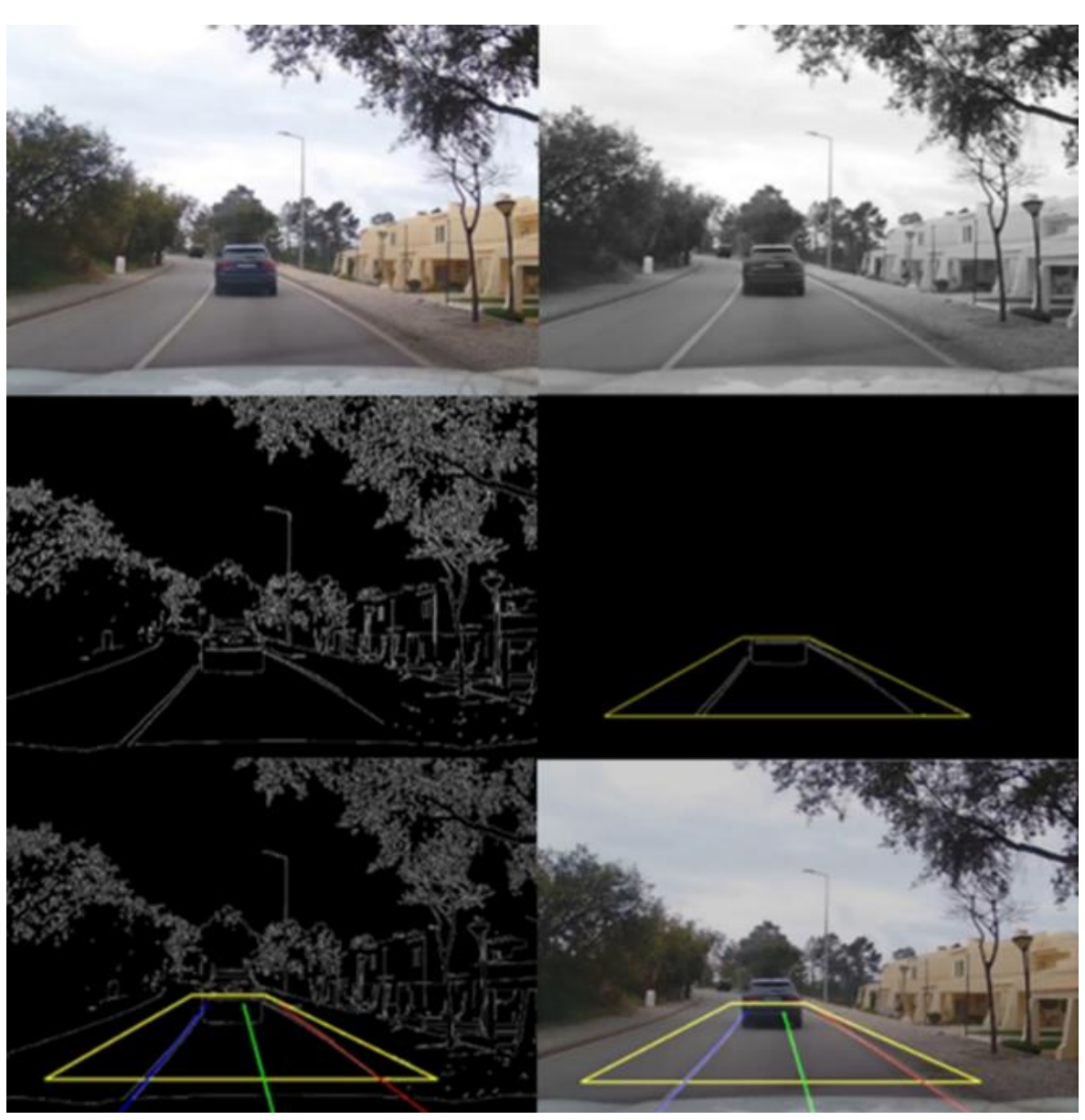

Fig. 2. Lane Detection

Fig 2. Illustrates the lane line detection process, presenting the frames from top to bottom and left to right. First, the original frame is converted to grayscale. Next, a blur is applied to improve the performance of Canny Edge Detection. Then an ROI is defined to filter out irrelevant information, allowing for the detection of lane lines. Finally, the detected lane lines are superimposed onto the original frame.

### E. Behaviour Analysis

Lateral displacement and oscillatory movements of detected vehicles were analyzed. *Lateral Movement:* The system calculates the deviation of the vehicle's centroid from the lane centre. *Oscillatory Behaviour:* Frequent sign changes in lateral movement indicate impaired driving [13], [14]. Metrics such as average lateral movement, lane offset, and sign changes were computed to classify driver behaviour. Alarms were triggered based on predefined thresholds. *Distracted Driver Alarm:* Activated when lateral movement exceeded a threshold (e.g., 0.3 pixels) and the vehicle deviated significantly from the lane centre (e.g., >40 pixels) [13]. *Impaired Driver Alarm:* Triggered when frequent oscillations (three or more sign changes) were detected, indicating erratic driving behaviour [13], [14]. Behavioural data for each detected object is logged into a CSV file, including frame number, object ID class (e.g., car, truck), lateral movement, lane offset and triggered alarms. Fig.3 illustrates the Flow chart of the Detection Algorithm.

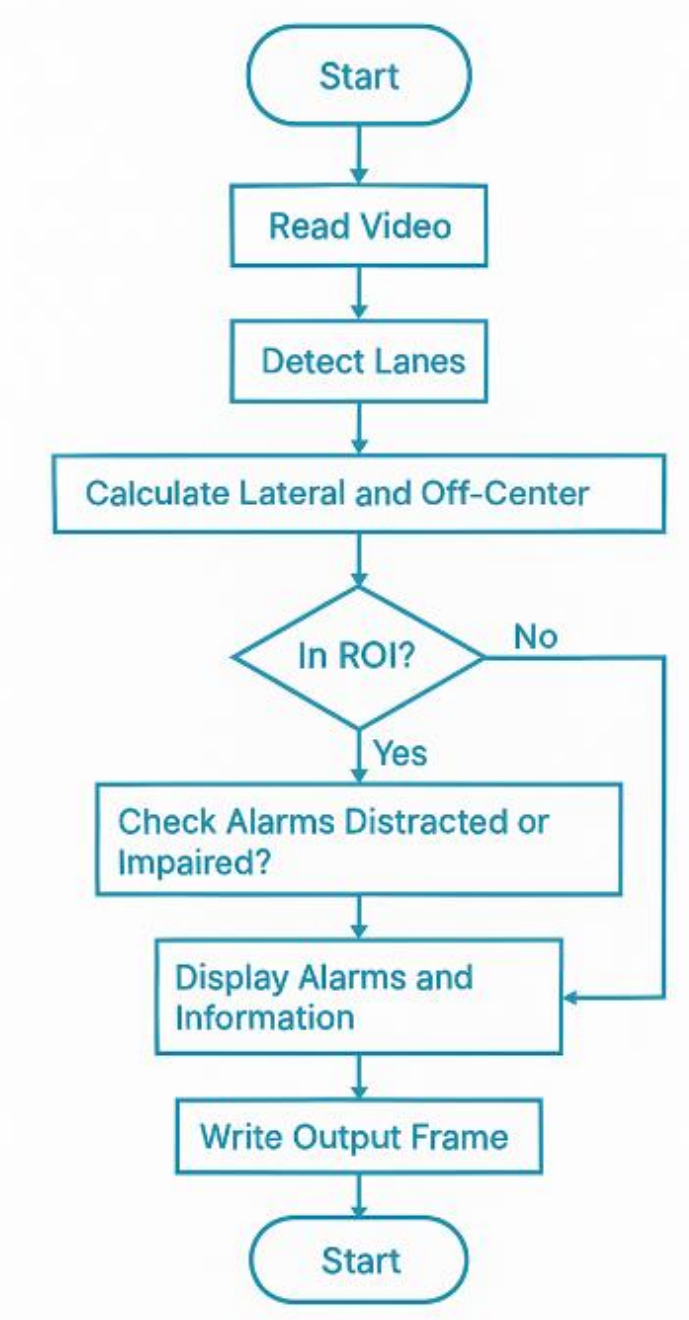

Fig. 3. Algorithm Flowchart

### F. Detection Frame

Fig. 4 illustrates the "Region of Interest" (ROI) used for lane line detection, with the detected area markings highlighted in yellow and the frame number in the bottom right corner. A blue and red overlay depicts the left and right lane lines. The Green line corresponds to the calculated centre of the lane. The yellow dot marks the centroid of the vehicle, and the green dot indicates the point on the lane centre closest to the vehicle's centroid. Annotations near the bounding box, including the object Class, the average lateral displacement of the object in pixels, the Lane off-centre (object distance from the lane centre in pixels), and the object direction, are displayed in blue. The Average Lateral Historical data is below the parameter data in blue text.

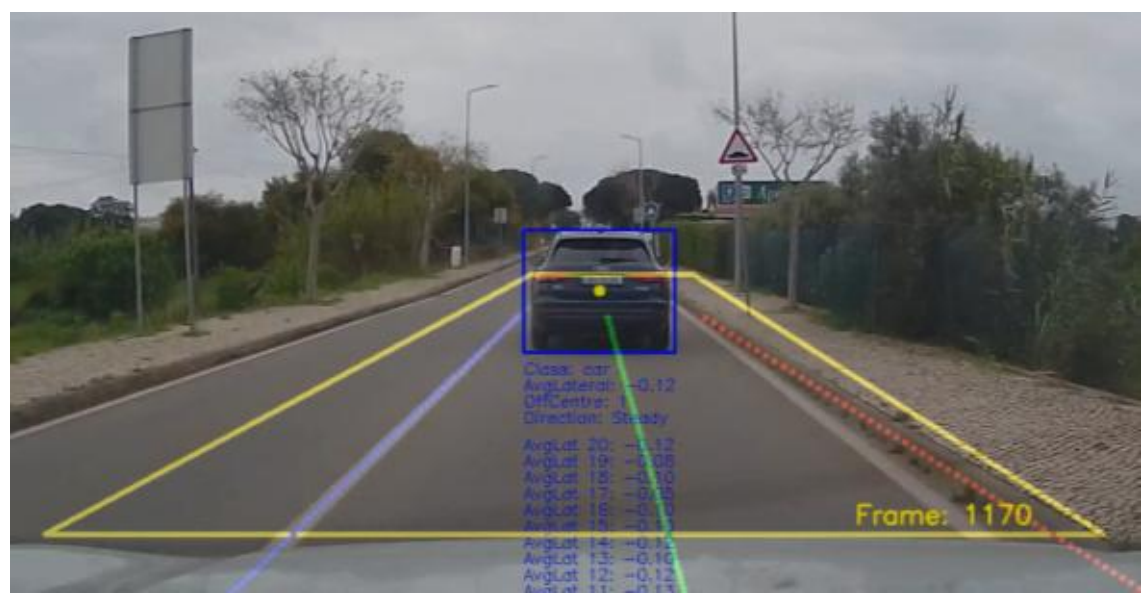

Fig. 4. Detection Frame

### G. Performance Evaluation

The system was tested on various driving scenarios to assess its accuracy and reliability. Challenges such as faded lane markings, complex road geometries, and environmental variability were addressed. Controlled simulations demonstrated the system's ability to detect unsafe driving behaviours effectively [13], [14].

## IV. Experimental Setup

The experimental setup was designed to evaluate the proposed driver behaviour classification system under diverse driving conditions. The setup involved controlled simulations and real-world testing to ensure the system's robustness and reliability. The ego vehicle was equipped with a Garmin 55 dashcam, centrally mounted on the windshield to ensure optimal field of view. The dashcam specifications included a Resolution of 1440p; Field of View: 122 degrees; Frame Rate: 30 FPS and data was stored using a micro SD card [12], [13]. The dashcam was powered via the vehicle's 12-volt outlet, and its placement was calibrated to maintain consistent alignment across different vehicles [15]. A driving route was planned to include various road types: 1) Highways: To test lane detection under clear and consistent markings; 2) Local Roads: To evaluate performance in urban environments with irregular markings. 3) Back Streets: To assess the system's adaptability to faded or absent lane markings [13], [14]. The route was driven at different times of the day to account for varying lighting conditions, including shadows, glare, and low-light scenarios [18]. To test the system's ability to detect distracted and impaired driving, controlled simulations were conducted. *Distracted Driving Simulation:* The target vehicle intentionally deviated from the lane centre to simulate distracted driving. This involved repeatedly crossing the lane boundary and returning to the lane centre. *Impaired Driving Simulation:* The target vehicle performed oscillatory movements, steering alternately left and right to simulate impaired driving behaviour. These movements were characterized by frequent lateral direction changes. All simulations were conducted in controlled environments to ensure safety, with no other vehicles or pedestrians present during the tests. Video recordings were captured during the simulations and real-world driving scenarios. The Garmin 55 dashcam provided high-resolution footage, which was processed using a custom Python-based detection algorithm. The system analyzed lateral displacement, lane positioning, and oscillatory behaviour to classify driver behaviour. The YOLOv8n object detection model was used to identify vehicles and other road users in real-time. Lane detection was performed using a multi-stage image preprocessing pipeline, including grayscale conversion, Gaussian blurring, adaptive thresholding, and Canny edge detection. Polynomial fitting via RANSAC regression was applied to estimate lane boundaries and the lane centre [18], [19], [16].

The system's performance was evaluated based on: *1) Detection Accuracy:* the ability to identify distracted and impaired driving behaviours. *2) False Positives:* Instances where everyday driving was incorrectly flagged as unsafe, and *3) Environmental Adaptability:* The system's robustness under varying road and lighting conditions. Behavioural data for each detected object was logged into a CSV file, including frame number, object ID, class, lateral movement, lane offset, and triggered alarms.

The experimental setup encountered a number of specific challenges, such as *Faded Lane Markings:* Reduced detection accuracy in areas with poorly maintained road markings. *Complex Road Geometries:* Difficulties detecting lane boundaries on curved roads and intersections. *Environmental Variability:* Adverse weather conditions, such as rain and fog, impacted detection reliability. These challenges highlight areas for future improvement, including integrating multi-modal sensors (e.g., LiDAR, radar) and expanding the dataset to include diverse driving scenarios [13], [14].

## V. RESULTS AND DISCUSSION

Results demonstrate the effectiveness of the proposed driver behaviour classification system in detecting distracted and impaired driving behaviours.

### *A. Detection of Distracted Driving*

The system successfully identified distracted driving behaviours based on lateral displacement and deviation from the lane centre. The "Distracted Driver Ahead" alarm was triggered when the average lateral movement exceeded the threshold of 0.3 pixels and the vehicle's deviation from the lane centre surpassed 40 pixels. The system consistently flagged distracted driving behaviours during controlled simulations and real-world tests [13], [14]. Annotated video frames displayed bounding boxes, lateral drift metrics, and triggered alarms, providing real-time feedback to the driver. The CSV logs confirmed the detection accuracy, with consistent alignment between visual indicators and recorded metrics [13].

Fig. 5 illustrates the detection of the Distracted Driver Ahead alarm, indicating that driver distraction has occurred. The frame analysis shows that the object has moved 56 pixels off the centre line, with an average lateral movement of 1.14 pixels. The average lateral movement computed over a history of frames is greater than the Average_Lateral_Movement Threshold, the object's horizontal deviation from the lane centre is more significant than the Off_Centre_Threshold, and the object is within the designated ROI, the "Distracted Driver Ahead" alarm is activated. The CSV file shows that Table I in frame 36, a "DISTRACTED DRIVER AHEAD" alarm is activated.

TABLE I. CSV FILE FOR FRAME 36

| **Frame** | **Monitored Variables** | | | | | | | | |
|---|---|---|---|---|---|---|---|---|---|
| | Object ID | Class | Cx | Cy | Avg. Lateral | Off-Centre | Direction | Sign Change | Alarms |
| 34 | 0 | car | 890 | 584 | 1.09 | 56 | Steady | 1 | DISTRACTED DRIVER AHEAD |
| 35 | 0 | car | 887 | 583 | 1.14 | 58 | Steady | 1 | DISTRACTED DRIVER AHEAD |
| **36** | **0** | **car** | **886** | **582** | **1.14** | **56** | **Steady** | **1** | **DISTRACTED DRIVER AHEAD** |
| 37 | 0 | car | 884 | 584 | 1.16 | 56 | Steady | 1 | DISTRACTED DRIVER AHEAD |
| 38 | 0 | car | 881 | 581 | 1.21 | 58 | Steady | 1 | DISTRACTED DRIVER AHEAD |

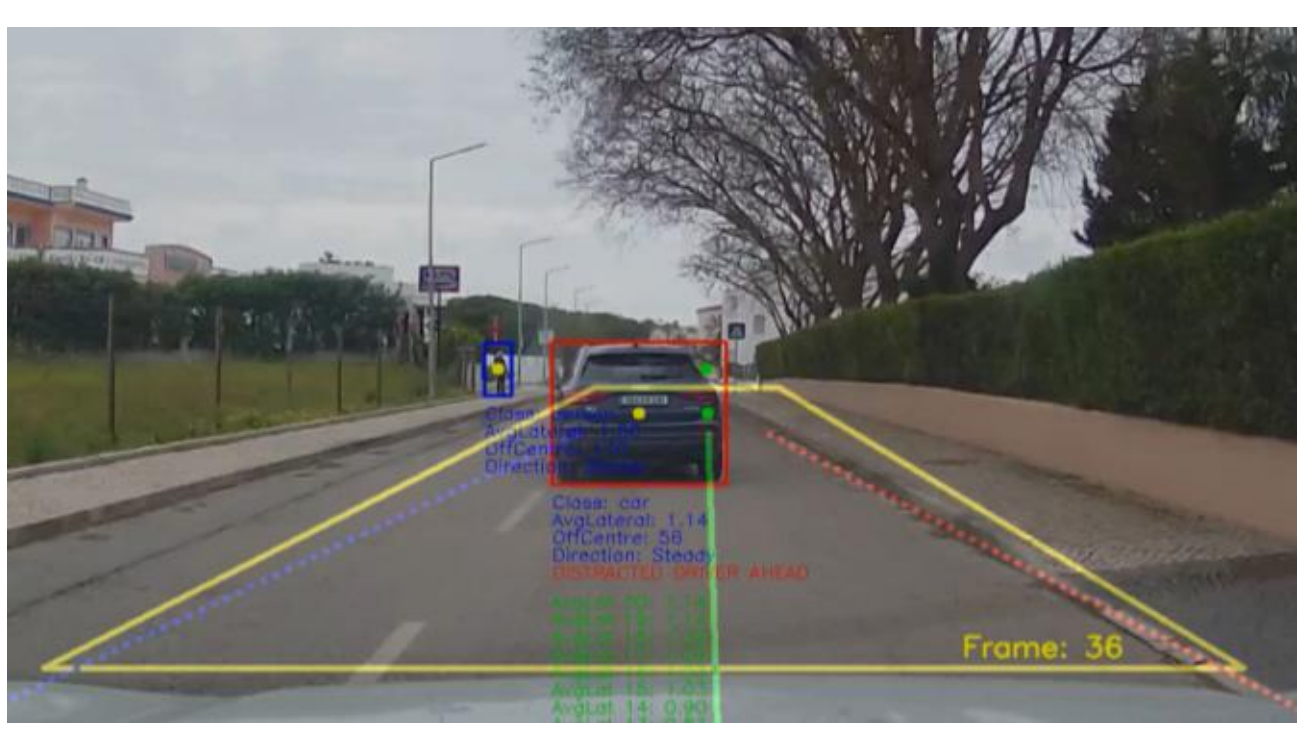

Fig. 5. Distracted Driver Alarm (Frame 36)

### *B. Detection of Impaired Driving*

The system effectively identified impaired driving behaviours characterized by oscillatory movements and frequent lateral direction changes. The "Impaired Driver Ahead" alarm was triggered when three or more sign changes in lateral movement were detected, indicating erratic driving behaviour. The lane centre is considered zero, and any movement over the lane line, depending on the Average_Lateral_Movement threshold limits, would constitute a sign change. Controlled simulations demonstrated the system's ability to detect impaired driving patterns reliably. The average lateral movement history was displayed with colour-coded severity levels (blue for safe, green for initial sign change, yellow for second sign change, and red for alarm activation). This visual representation enhanced the interpretability of the detection results. Fig. 6 illustrates that the algorithm has detected the first lateral movement with a sign change, which shows an oscillation. The average lateral historical displayed data has now been changed to green.

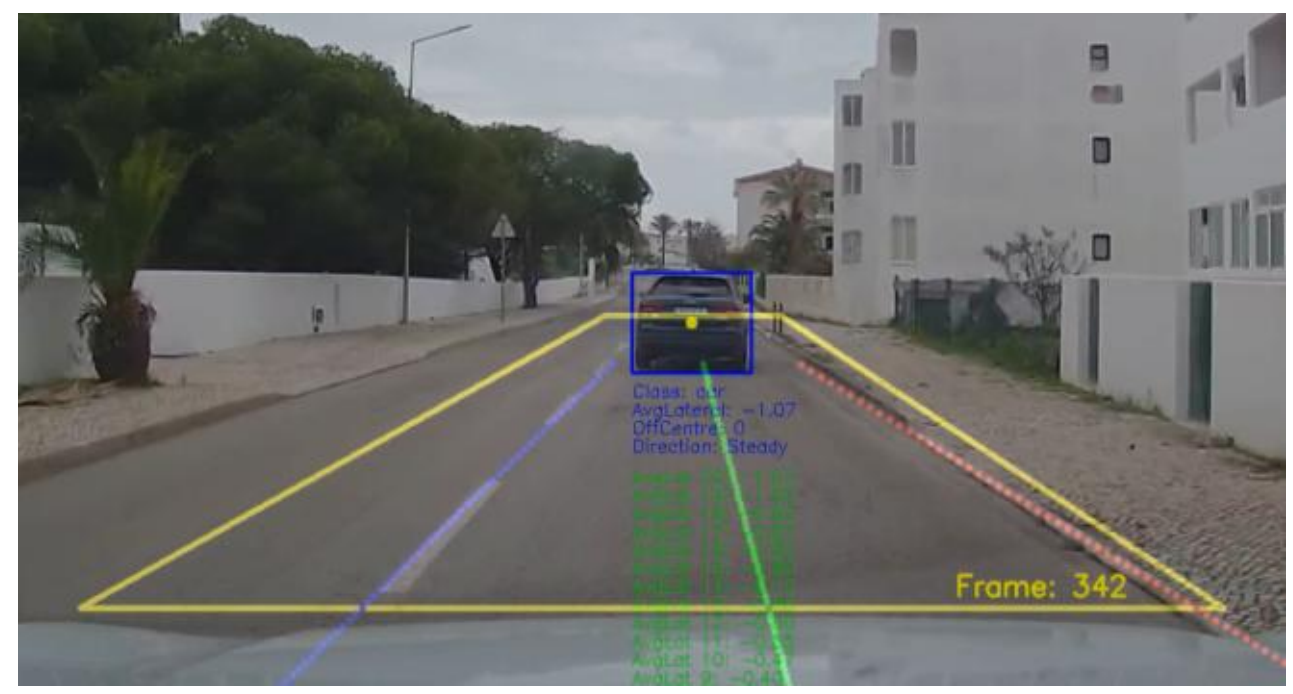

Fig. 6. The First Lateral Sign Change (Frame 342)

Fig. 7 illustrates that the algorithm has detected the second lateral movement, where a sign change shows an oscillation. The average lateral historical displayed data has now been changed to yellow.

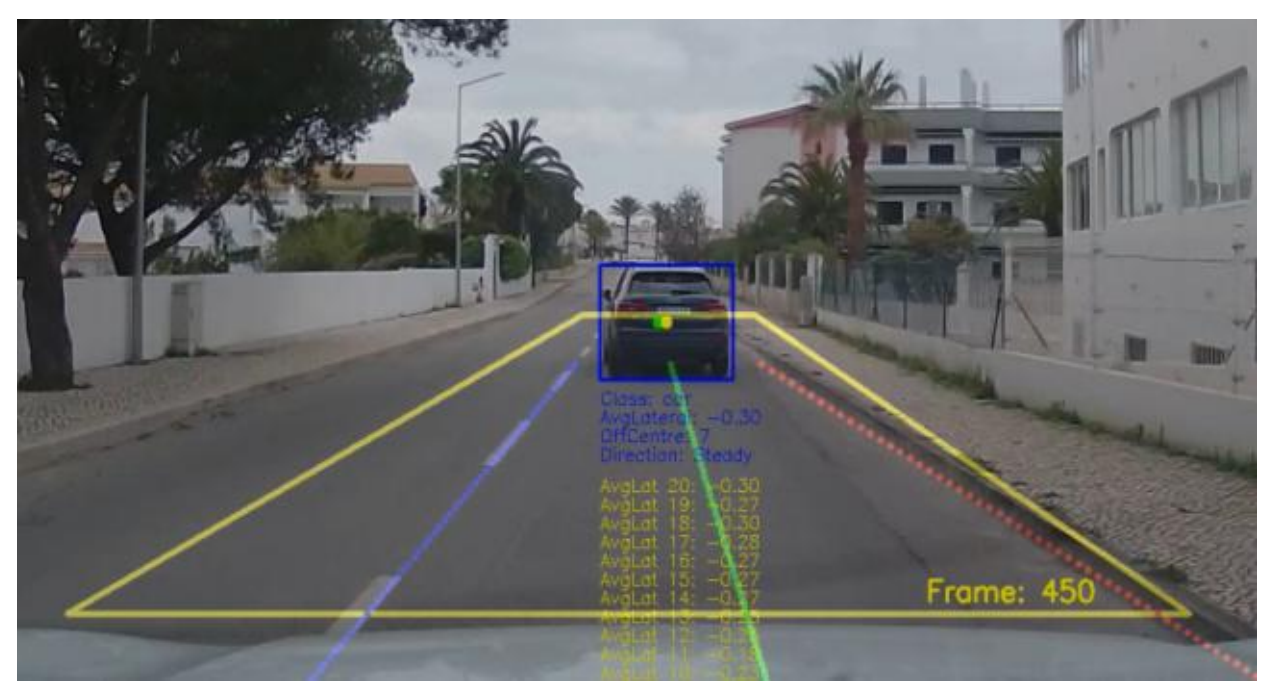

Fig. 7. The Second Lateral Sign Change (Frame 450)

Fig. 8 illustrates that the algorithm must detect the third lateral movement where a sign change shows an oscillation. Now, the "Impaired Driver Ahead" alarm has been activated. The average lateral historical displayed data has now been changed to red.

TABLE II. CSV FILE FOR FRAME 483

| Frame | Monitored Variables | | | | | | | | |
|---|---|---|---|---|---|---|---|---|---|
| | *Object ID* | *Class* | *Cx* | *Cy* | *Avg. Lateral* | *Off-Centre* | *Direction* | *Sign Change* | *Alarms* |
| 481 | 0 | car | 910 | 573 | 0.17 | 6 | Steady | 3 | IMPAIRED DRIVER AHEAD |
| 482 | 0 | car | 911 | 572 | 0.22 | 4 | Steady | 3 | IMPAIRED DRIVER AHEAD |
| **483** | **0** | **car** | **912** | **573** | **0.2** | **4** | **Steady** | **3** | **IMPAIRED DRIVER AHEAD** |
| 484 | 0 | car | 912 | 573 | 0.2 | 5 | Steady | 3 | IMPAIRED DRIVER AHEAD |
| 485 | 0 | car | 915 | 579 | 0.2 | 3 | Steady | 3 | IMPAIRED DRIVER AHEAD |

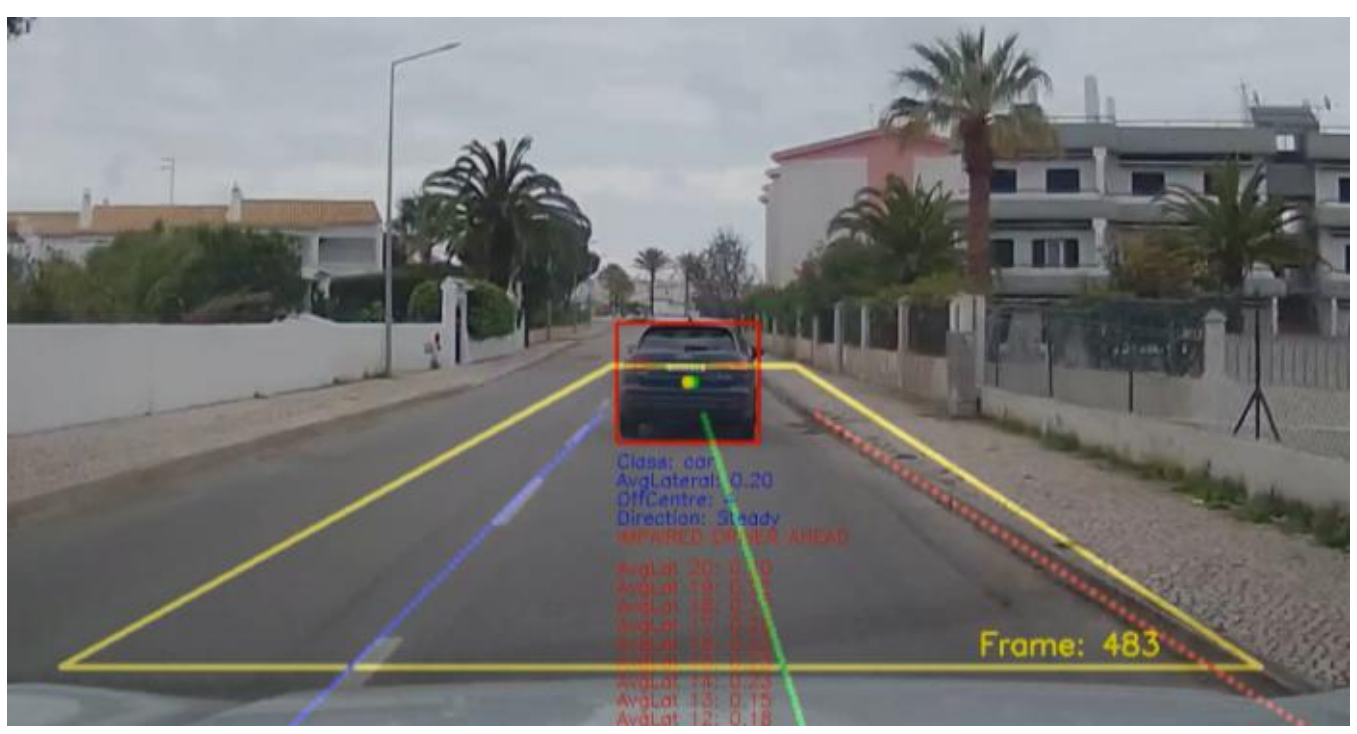

Fig. 8. The Third Lateral Sign Changes, and the "Impaired Driver Ahead Alarm" is Activated (Frame 483)

Table II in frame 483, shows three sign changes, and the "IMPAIRED DRIVER AHEAD" alarm is activated.

*C. Challenges:*

Adverse weather conditions, such as rain and fog, impacted the reliability of oscillatory movement detection. The system occasionally misclassified lateral movements caused by road conditions (e.g., potholes) as impaired driving behaviours.

*D. Quantitative Results*

The system's performance was evaluated using detection accuracy, false positives, and environmental adaptability metrics.

*1) Distracted Driving:* The system accurately detected lateral deviations, with minimal false positives under ideal road conditions. However, accuracy decreased in scenarios with faded lane markings or complex road geometries.

*2) Impaired Driving:* The system reliably detected oscillatory movements, with consistent alarm activation during controlled simulations. False positives were observed in scenarios with uneven road surfaces.

The CSV logs provided a detailed record of object-level data, including Frame Number, Object ID, Class, Lateral Movement, Lane Offset, and triggered Alarms. Cross-validation between the annotated video frames and CSV logs confirmed the system's reliability in detecting unsafe driving behaviours.

Results highlight the system's potential to enhance road safety by detecting distracted and impaired driving behaviours. Integrating advanced computer vision techniques, such as YOLO-based object detection and RANSAC regression for lane fitting, contributed to the system's robustness and adaptability [19], [20]. The system provided timely feedback, enabling proactive responses to unsafe driving behaviours. The monocular camera and lightweight YOLOv8n model ensure compatibility with non-autonomous vehicles, making the system suitable for mass deployment [15].

*Limitations:* Detection accuracy was compromised under adverse weather conditions and in areas with faded lane markings [18]. Real-time performance degraded when processing scenes with multiple moving objects, highlighting the need for hardware acceleration [13].

### E. Future Improvements

Future work will focus on:

*1) Multi-Modal Sensors: Integrating LiDAR, RADAR, and infrared sensors to improve detection accuracy under adverse conditions [18].*

*2) Dataset Expansion: Including diverse driving scenarios, such as extreme weather and urban traffic, to enhance algorithmic robustness [13].*

*3) False Positive Reduction: Developing mechanisms to distinguish between road-induced movements and unsafe driving behaviours.*

## VI. Conclusion

This research demonstrates the viability of external observation techniques for classifying driver behaviour, focusing on detecting distracted and impaired driving. By employing advanced computer vision methodologies, including object tracking, lateral displacement analysis, and lane position monitoring, the system establishes a robust framework for identifying hazardous driving patterns. Integrating the YOLO object detection model and custom lane estimation algorithms enables real-time hazard detection without relying on inter-vehicular communication. It is particularly suitable for environments where vehicles lack V2X connectivity features.

Experimental results highlight the system's ability to accurately trigger alarms for unsafe driving behaviours, such as excessive lateral movement and oscillatory patterns. This highlights its potential to enhance road safety between manually driven and fully autonomous vehicles during the transitional phase. The system's adaptability across diverse road and environmental conditions further underscores its practical applicability. The proposed methodology represents a significant step forward in driver behaviour classification, offering a scalable and effective solution for improving road safety. By addressing its limitations and incorporating future advancements, the system has the potential to play a critical role in the safe coexistence of human-driven and autonomous vehicles, ultimately contributing to the broader vision of fully autonomous transportation systems.